\documentclass[11pt]{article}

\usepackage[preprint]{acl}
\usepackage{booktabs}
\usepackage{multirow}
\usepackage{times}
\usepackage{latexsym}
\usepackage{algorithm}
\usepackage{algorithmic}

\usepackage{amsmath}
\usepackage{amssymb}

\usepackage[T1]{fontenc}

\usepackage[utf8]{inputenc}

\usepackage{microtype}
\usepackage{enumitem} 
\usepackage{inconsolata}
\usepackage[most]{tcolorbox}
\usepackage{graphicx}
\usepackage{amssymb}
\usepackage{microtype}
\usepackage{amsfonts}
\usepackage{xcolor}
\definecolor{lavender}{RGB}{230,230,250} 
\definecolor{softlavender}{RGB}{238, 223, 255} 
\definecolor{selfevolagent}{RGB}{220,211,237} 
\definecolor{selfevolagent_dark}{HTML}{245b9c} 
\definecolor{selfevolagent_light}{HTML}{6093d0}
\definecolor{selfevolagent_lighter}{HTML}{6093d0}

\title{ForgetBench: Benchmarking Forgetting Dynamics of Long-Term Parametric Memory in Language Models}

\author{
Ruxi Gu$^{1,2}$ \quad Zhenliang Zhang$^{2}$ \quad Zilei Wang$^{1*}$ \quad Wei Wang$^{2,3}$\thanks{Corresponding authors. Emails: \texttt{zlwang@ustc.edu.cn}, \texttt{wangwei@ustb.edu.cn}.}\\
$^{1}$Department of Automation, University of Science and Technology of China\\
$^{2}$State Key Laboratory of General Artificial Intelligence, BIGAI\\
$^{3}$School of Computer and Communication Engineering, \\ University of Science and Technology Beijing
}

\begin{document}
\maketitle
\begin{abstract}
Large language models (LLMs) have demonstrated strong capabilities in knowledge acquisition and reasoning, yet their ability to retain previously acquired knowledge under repeated updates remains insufficiently understood. Existing evaluation paradigms primarily focus on single-step reasoning or static knowledge editing, which fail to capture the temporal dynamics of knowledge retention and degradation during continual model modification.  
In this work, we propose \textbf{ForgetBench}, a benchmark for measuring forgetting dynamics in the long-term parametric memory of LLMs under continual knowledge editing. 
ForgetBench introduces two complementary evaluation paradigms, namely concept-based QA and scenario-based QA, to probe the retention of isolated facts and structured relational knowledge.
Building upon a sequential editing framework, we construct temporally ordered knowledge streams and evaluate model behavior across multiple editing stages.  
To quantify these dynamics, we introduce a unified evaluation framework that indexes memory age by the number of subsequent edits and measures temporal decay, retention strength, and cross-instance stability. 
Extensive experiments across diverse models and editing methods reveal distinct forgetting trajectories and a recurring tension between long-term retention and generalization quality.
Our findings highlight the need for more robust memory mechanisms that can effectively acquire, update, and preserve knowledge over time in future LLMs.
\end{abstract}

\section{Introduction}

Memory is essential for intelligent systems operating in dynamic environments, where knowledge must be continuously acquired, retained, and updated over time. Inspired by cognitive studies of memory \cite{atkinson1968human}, memory is commonly divided into working memory and long-term memory. While working memory supports transient information manipulation within a single interaction, long-term memory enables durable knowledge retention and gradual forgetting across extended periods. For large language models (LLMs), understanding such long-term memory dynamics is increasingly important as models are expected to adapt continuously after deployment.

Despite the rapid progress of LLMs, most existing evaluation paradigms \cite{yang2018hotpotqa, bai2024longbench} primarily focus on working memory abilities, such as in-context reasoning over long inputs or multi-hop question answering within a single inference process, rather than persistent knowledge retention after updates. Recent studies have begun to investigate long-term memory in interactive systems, including multi-session conversational memory \cite{wu2024longmemeval} and evolving agent memory evaluation \cite{ding2026memground}. However, these benchmarks mainly evaluate externally maintained memory, where retention depends on retrieval mechanisms, memory organization, or interaction histories. 
In contrast, our work focuses on intrinsic parametric memory, where knowledge is encoded in model parameters and its accessibility evolves as subsequent parameter updates accumulate.
Such a setting is markedly important for continual adaptation, where models repeatedly acquire new knowledge through parameter modification.

Knowledge editing \cite{meng2022mass, fang2024alphaedit} has recently emerged as a promising paradigm for updating factual knowledge in LLMs without full retraining. 
Existing editing benchmarks, including zsRE \cite{levy2017zero}, CounterFact \cite{meng2022locating}, UnKE \cite{deng2024everything}, and WikiBigEdit \cite{thede2025wikibigeditunderstandinglimitslifelong}, assess edit accuracy, generalization, locality, and performance under repeated updates. However, their aggregate scores provide limited visibility into how individual edits evolve as later updates accumulate.
Consequently, they cannot answer a fundamental question for reliable lifelong adaptation: whether knowledge acquired today can survive future modifications.

To address this gap, we propose \textbf{ForgetBench}, a benchmark for studying forgetting dynamics in the long-term parametric memory of LLMs under continual knowledge editing. Rather than reporting editing performance at a single stage, ForgetBench repeatedly revisits historical edits as subsequent updates accumulate and defines their memory age by the number of intervening edits. We construct two complementary settings: (i) concept-based QA, which isolates atomic factual updates for controlled analysis of parameter interference, and (ii) scenario-based QA, which introduces structured relational knowledge through multi-agent interaction graphs. 
By tracking model performance across editing steps, ForgetBench converts historical knowledge retention into age-conditioned forgetting curves, revealing whether edited memories persist, degrade, or fluctuate over time.

Our main contributions can be summarized as follows: (1) We introduce ForgetBench, a unified benchmark designed to evaluate the long-term parametric forgetting dynamics of LLMs under continual knowledge editing. Moving beyond static, single-step edit evaluations, ForgetBench enables temporal analysis of knowledge acquisition, retention, and forgetting as parametric memories age. (2) We design two complementary QA construction paradigms that capture both isolated factual updates and structured relational knowledge. Based on temporally ordered knowledge streams, we develop an evaluation framework to quantify retention strength, temporal decay, and cross-instance stability. (3) Extensive experiments across multiple LLMs and editing methods demonstrate that existing editing approaches face challenges in maintaining both long-term factual retention and generalization capability. Our findings provide a systematic analysis of the temporal limitations of current knowledge editing paradigms and motivate future research toward more robust memory mechanisms.

\section{Related Works}

\subsection{Working \& Long-Term Memory}

Memory in cognitive science is commonly divided into short-term and long-term components. Following the Atkinson-Shiffrin memory model \cite{atkinson1968human}, short-term memory supports temporary information manipulation during ongoing processing, whereas long-term memory enables durable storage, accumulation, and retrieval of knowledge over extended periods. This distinction has inspired recent efforts to investigate whether language models can maintain information beyond a single inference context.

Early studies on LLM memory have predominantly focused on working memory capabilities, including long-context reasoning, context engineering \cite{asai2023self}, key-value reuse and compression \cite{chevalier2023adapting, li2024snapkv}, and chain-of-thought reasoning \cite{wei2022chain, xu2025softcot}. These approaches improve the utilization of information temporarily available in the context window, but they do not examine whether newly acquired knowledge can be persistently stored in model parameters. Similarly, retrieval-augmented generation (RAG) and external memory mechanisms \cite{lewis2020retrieval} enhance knowledge access through external retrieval, where memory retention depends on storage, retrieval, or management strategies rather than intrinsic parameter updates. 

Recent benchmarks have extended memory evaluation toward long-term interactive scenarios. Existing datasets such as HotpotQA \cite{yang2018hotpotqa}, LongBench \cite{bai2024longbench}, and LoCoMo \cite{maharana2024evaluating} evaluate reasoning over complex contexts or conversational history, while newer benchmarks and studies further investigate memory consistency in multi-session \cite{wu2024longmemeval, hu2026evermembench, ding2026memground} and evolving agent environments \cite{uddin2026recall}. These studies provide important insights into how LLM-based systems retrieve and organize information over time. However, their memory evolution is primarily governed by external memory modules, retrieval policies, or interaction histories, and therefore the observed forgetting reflects failures in memory access or management rather than interference among internal model parameters.

The forgetting curve provides another perspective for characterizing memory degradation. Inspired by Ebbinghaus' forgetting curve \cite{ebbinghaus2013image}, Liu et al.~\cite{liu2024forgetting} introduce forgetting curves to analyze memorization in long-context language models. However, their formulation defines memory age according to contextual distance, measuring the decay of information presented within an input sequence. 
In contrast, ForgetBench focuses on intrinsic parametric memory under continual knowledge editing. It defines memory age by the number of subsequent editing operations and repeatedly evaluates earlier edits along this dimension. This formulation exposes long-term stability, forgetting dynamics, and changes in knowledge accessibility that are not visible from a single editing outcome.

\subsection{Knowledge Editing}

Knowledge editing aims to modify factual knowledge in large language models without full retraining. Early methods based on causal intervention and direct weight modification \cite{mitchell2021fast} showed that factual associations can be localized and explicitly altered. Subsequent work improved editing scale and locality through methods such as MEMIT \cite{meng2022mass}, AlphaEdit \cite{fang2024alphaedit}, hypernetwork-based editors \cite{tan2023massive,li2025reinforced,gu2026hierarchical}, and locate-then-edit approaches \cite{pan2025precise}. More recent lifelong editing methods, including LightEdit \cite{jung2026scalablelifelongknowledgeediting} and LPD-Edit \cite{LIU2026123848}, further target scalability and stability under sequential updates. 
However, these methods primarily focus on how to perform or stabilize edits, while the age-dependent evolution of individual edited knowledge items remains less explicitly characterized.

Editing benchmarks measure edit success, paraphrase generalization, locality, and performance over repeated update streams \cite{levy2017zero,meng2022locating,deng2024everything,thede2025wikibigeditunderstandinglimitslifelong}. These evaluations establish whether editing remains effective under different settings, but provide limited characterization of the forgetting trajectories of individual edits as their memory age increases. ForgetBench complements these efforts by repeatedly evaluating historical edits across editing stages, enabling explicit analysis of retention, temporal decay, and forgetting dynamics.

\subsection{Overview of ForgetBench}

\begin{figure}[t]
\centering
\includegraphics[width=0.45\textwidth]{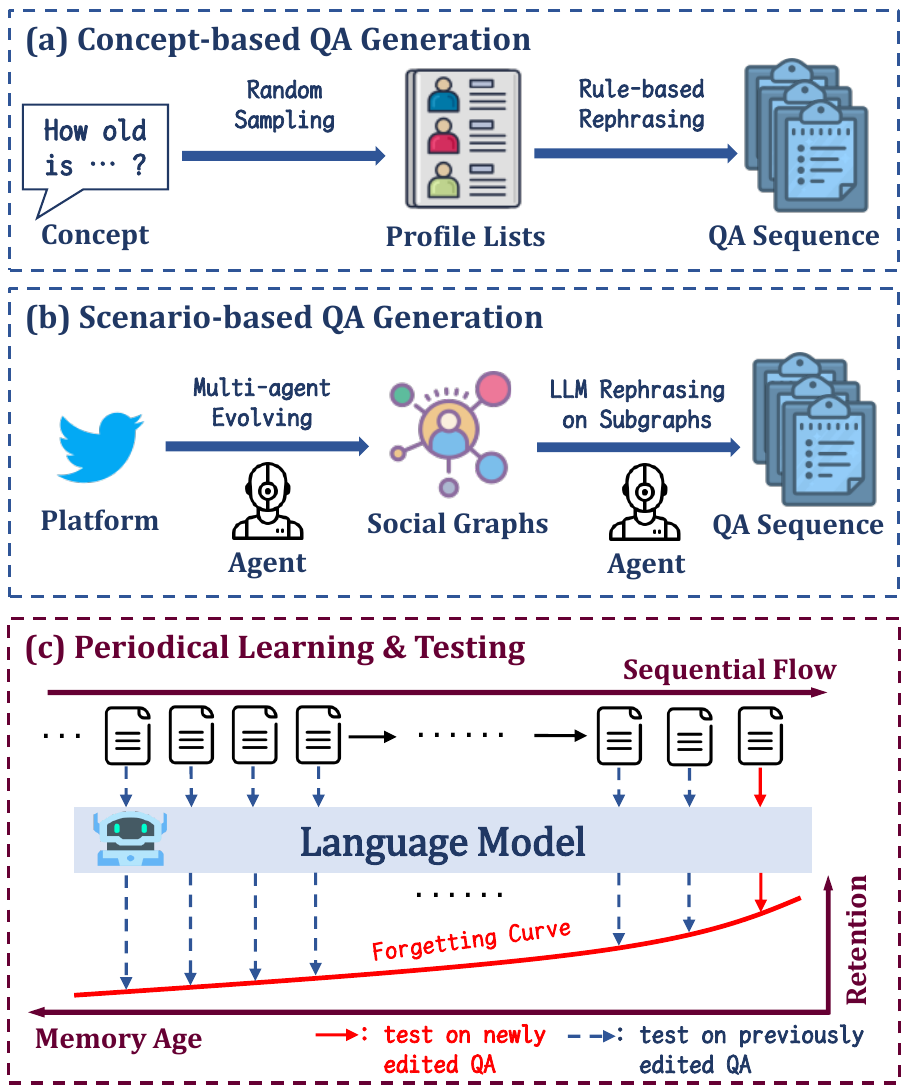} 
\caption{Overview of ForgetBench. We construct two categories of QA sequences. Concept-based QA (a) focuses on isolated concepts, while scenario-based QA (b) evaluates structured knowledge understanding. Subsequently, sequential editing (c) exposes the model to these QA sequences and evaluates forgetting dynamics across editing stages.}
\label{framework}
\end{figure}

As illustrated in Fig.~\ref{framework}, ForgetBench is composed of two complementary QA construction paradigms, namely \textbf{concept-based QA} and \textbf{scenario-based QA}, which differ in both the granularity and structural organization of knowledge.

The concept-based setting focuses on atomic factual attributes, where each subject is associated with a single property and independently perturbed over time. This yields disentangled instances for studying the retention of localized factual updates. In contrast, the scenario-based setting constructs QA pairs over structured relational knowledge induced by an interaction graph. A knowledge graph is first generated through simulated agent-item interactions, and QA instances are then derived from sampled subgraphs and transformed into natural language contexts, requiring reasoning over interconnected entities. Given these two complementary constructions, ForgetBench evaluates models under a sequential editing paradigm, where new knowledge is continuously injected over time. 
By periodically testing previously and newly edited instances, we obtain fine-grained trajectories that connect recent editing performance with the long-term accessibility of parametric memories.

\subsection{Data Generation}

\noindent \textbf{Concept-based QA.} The concept-based setting constructs independent QA instances over atomic subjects. Each subject is assigned an attribute, whose pre-edit value is sampled from a predefined distribution and then replaced by an independently sampled post-edit value as the editing target. Each QA instance follows a fixed attribute query template $Q$, with the post-edit attribute $A^{right}$ serving as the ground-truth answer and the pre-edit value $A^{wrong}$ representing the overwritten memory (e.g., $Q$: \textit{``How old is Donald Rodriguez?''}, $A^{right}$: \textit{``22''}, $A^{wrong}$: \textit{``79''}). Since all subjects are sampled independently, this setting removes inter-instance dependencies and provides a controlled environment for analyzing localized knowledge updates.

\noindent \textbf{Scenario-based QA.} The scenario-based setting constructs QA pairs over structured relational knowledge induced by a dynamic interaction graph. We initialize a set of agents $\mathcal{A}$ and evolve their interactions through multiple simulation rounds, where active agents participate in collaborative behaviors that generate entities and relational edges. This process yields a heterogeneous knowledge graph $\mathcal{G}$ consisting of agents, items, and their interactions. The detailed algorithm pseudocode is provided in the supplementary material.

After graph construction, we sample multiple subgraphs with controlled overlap to maintain structural diversity and convert each into a natural-language context through rule-based rephrasing. An LLM uses these contexts to generate multiple-choice QA pairs with one correct answer and plausible distractors drawn from the subgraph. The generated context is used during question construction; editing and evaluation use the resulting QA instances. This protocol tests whether relational knowledge acquired through model updates supports subsequent question answering.

\begin{figure}[htbp]
  \centering
  \includegraphics[width=0.95\linewidth]{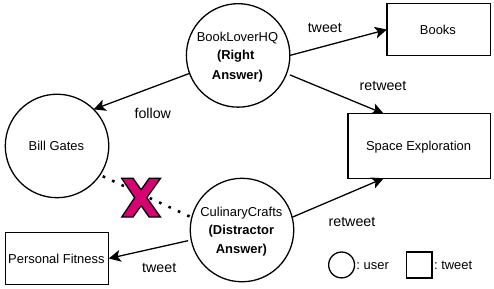}
  \caption{An example subgraph used for QA generation.}
  \label{fig_graph1}
\end{figure}

As illustrated in Fig.~\ref{fig_graph1}, a query $Q$ tests multi-hop relational reasoning: \textit{``Which user followed BillGates and retweeted a post about space exploration? A. BookLoverHQ B. FoodieFeest C. TravelBug101 D. CulinaryCrafts''}. The correct answer $A^{right}$ (\textit{A. BookLoverHQ}) satisfies both paths, while $A^{wrong}$ (\textit{D. CulinaryCrafts}) acts as a distractor lacking the following link to BillGates.

\begin{figure}[htbp]
  \centering
    \includegraphics[width=0.95\linewidth]{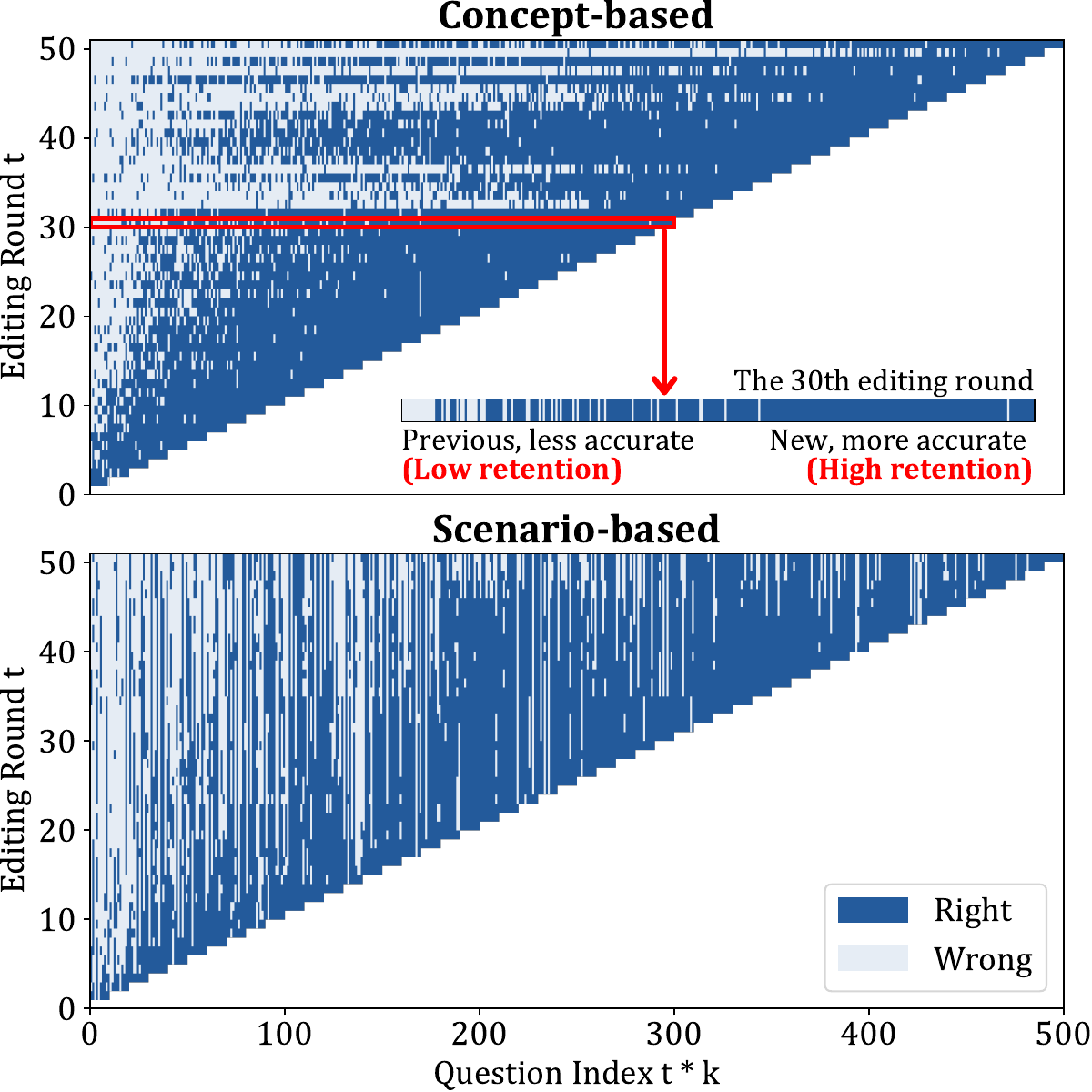}
  
  \caption{The evaluation matrices $\mathcal{Z}$ of Concept-based (upper) and Scenario-based (lower) tests.}
  \label{fig14}
\end{figure}

\section{Forgetting Quantification}

To comprehensively evaluate long-term memory across different temporal horizons, we examine not only queries within a single sequential editing process, but also variations in the overall QA sequence length. Specifically, given a maximum sequence length $T$ and evaluation interval $k$, we construct a set of sequence lengths $\{t_n\}_{n=0}^{T/k}$, where each $t_n$ corresponds to $n \cdot k$ editing steps. 
At each editing checkpoint, all previously introduced knowledge items are evaluated to measure how their accessibility changes over subsequent updates. 
We denote by $\mathcal{Z}[t, i]$ the correctness of the model on the query corresponding to the $i$-th edit, evaluated after the $t$-th evaluation round, where $i \leq t \cdot k$. Each entry $\mathcal{Z}[t, i] \in \{0,1\}$ indicates whether the knowledge item is correctly recalled. 

Fig.~\ref{fig14} visualizes the evaluation matrix $\mathcal{Z}$ under both settings, revealing distinct spatiotemporal structures. In the concept-based setting, correctness is highly dispersed across both edits and evaluation rounds, reflecting independently introduced factual updates that are uniformly vulnerable to interference. 
In contrast, the scenario-based setting exhibits stronger vertical coherence, where correctness often persists across successive rounds, indicating more structured recall trajectories under relational queries.

Overall, our evaluation framework characterizes continual knowledge editing through a unified spatiotemporal perspective over $\mathcal{Z}$, where the horizontal axis corresponds to memory age (i.e., forgetting dynamics over $\Delta = tk - i$), and the vertical axis tracks individual knowledge instances across successive editing rounds. Based on this formulation, we construct a set of complementary metrics that capture three key dimensions: (1) pointwise editing correctness, (2) temporal decay along memory age, and (3) cross-instance stability across the editing trajectory. Together, these metrics provide a unified characterization of both global performance trends and fine-grained memory evolution dynamics.

\subsection{Pointwise Editing Performance} 

We summarize editing performance at the level of individual knowledge updates, considering both the accuracy of recent edits and the accessibility of previously introduced knowledge.
Edit Success (ES) measures whether newly edited knowledge is correctly acquired at the time of insertion. It is defined as the average performance on the diagonal entries of $\mathcal{Z}$, Retention (Ret) evaluates how well previously edited knowledge is preserved over time. It is defined as the average performance over all past knowledge. ES and Ret are computed as:
\begin{equation}
    \text{ES} = \frac{k}{T}\sum_{t=1}^{T/k} \mathcal{Z}_{t, t k}, 
    \text{Ret} = \frac{k}{T}\sum_{t=1}^{T/k} (\frac{1}{t k - 1}\sum_{i=1}^{t k - 1} \mathcal{Z}_{t,i}).
\end{equation}


ES summarizes accuracy on recent edits at evaluation checkpoints, whereas Ret summarizes accuracy on historical edits accumulated up to those checkpoints.

\subsection{Temporal Decay Dynamics} 

To characterize the dynamic decay of long-term memory over time, we introduce the Forgetting Curve to quantify the average retention of knowledge after undergoing successive editing operations. Unlike ES or Ret that merely compute the overall average accuracy, the Forgetting Curve explicitly models the relationship between memory age and knowledge retention. For an arbitrary memory age $\Delta$, we consider all knowledge states satisfying $\Delta=tk-i$, namely, all knowledge items that have experienced $\Delta$ subsequent edits. Based on this formulation, the Forgetting Curve is defined as
\begin{equation}
    \mathcal{F}(\Delta)=\mathbb{E}[\mathcal{Z}[t,i]\mid tk - i = \Delta].
\end{equation}

$\mathcal{F}(\Delta)$ yields an age-conditioned trajectory of long-term parametric memory under continual editing. As $\Delta$ increases, the curve reveals whether historical knowledge decays, remains stable, or changes non-monotonically.

Furthermore, we define the memory half-life based on the Forgetting Curve, which characterizes the number of editing steps required for memory retention to decay to half of its initial level. Given the Forgetting Curve $\mathcal{F}(\Delta)$, we define the initial retention as $\mathcal{F}(0)$ and the half-life threshold as $\theta=\frac{1}{2}\mathcal{F}(0)$. The memory half-life $\Delta_{1/2}$ is then defined as the smallest $\Delta$ such that $\mathcal{F}(\Delta)\le \theta$, i.e., the first crossing point of the curve with the threshold.

To improve robustness against local fluctuations in $\mathcal{F(\Delta)}$, we apply a lightweight smoothing preprocessing step to reduce small-scale noise. Since the curve is evaluated at discrete steps, the exact crossing point may lie between two adjacent samples. We therefore apply linear interpolation: if $\mathcal{F}(\Delta_{k-1})>\theta$ and $\mathcal{F}(\Delta_k)\le \theta$, the half-life is estimated as
\begin{equation}
    \Delta_{1/2} = \Delta_{k-1} + \frac{ \mathcal{F}(\Delta_{k-1})-\theta }{ \mathcal{F}(\Delta_{k-1})-\mathcal{F}(\Delta_k) } (\Delta_k-\Delta_{k-1}).
\end{equation}

We analyze $\mathcal{F}(\Delta)$ through its average accuracy and its direction across memory ages. The Area Under the Forgetting Curve (AUF) measures the mean accuracy maintained over the observed editing horizon, while Decay summarizes signed changes between adjacent ages and reveals the overall direction of the age-conditioned trajectory:
\begin{equation}\text{AUF}=\mathbb{E}[\mathcal{F}(\Delta)],\text{Decay}=\mathbb{E}[\mathcal{F}(\Delta)-\mathcal{F}(\Delta+1)].\end{equation} 
A larger AUF reflects stronger accuracy throughout the editing horizon. Positive Decay indicates an overall decrease as memory age grows, whereas negative Decay indicates an overall increase. This signed measure captures the trajectory's direction.

\subsection{Cross-Instance Stability} 

While the Forgetting Curve aggregates responses by memory age, Temporal Consistency (TC) follows the evolution of each edited item across successive evaluation rounds. For item $i$, let $\mathcal{P}_i$ denote the set of consecutive available observations after removing unavailable states. We define the number of acquired states and the number of subsequent forgetting events as $N_{\mathrm{acc}}=\sum_i\sum_{t\in\mathcal{P}_i}\mathbf{1}[\mathcal{Z}[t,i]=1]$ and $N_{\mathrm{forget}}=\sum_i\sum_{t\in\mathcal{P}_i}\mathbf{1}[\mathcal{Z}[t,i]=1\land\mathcal{Z}[t+1,i]=0]$, respectively. 
\begin{equation}\mathrm{TC}=1-\frac{N_{\mathrm{forget}}}{N_{\mathrm{acc}}}.\end{equation} 
Items with fewer than two available observations contribute no transition, and TC is undefined when no correct starting state is observed. The metric conditions on a correct response at the current evaluation round and measures whether that response survives the next editing interval. Thus, transitions from an incorrect state are excluded because they describe acquisition rather than forgetting, while correct-to-correct transitions increase TC and correct-to-incorrect transitions decrease it. 

Higher TC indicates that correctly accessible edited knowledge persists across more subsequent updates. ES, Ret, and TC therefore describe three related but distinct aspects of continual editing: ES measures the accuracy of newly edited items at evaluation checkpoints, Ret summarizes accuracy over previously edited items, and TC measures the conditional persistence of correct responses along their trajectories. Ret determines the overall level of historical accuracy, whereas TC characterizes how reliably that level is maintained from one evaluation round to the next.

\begin{figure*}[bt]
  \centering
    \centering
    \includegraphics[width=1\linewidth]{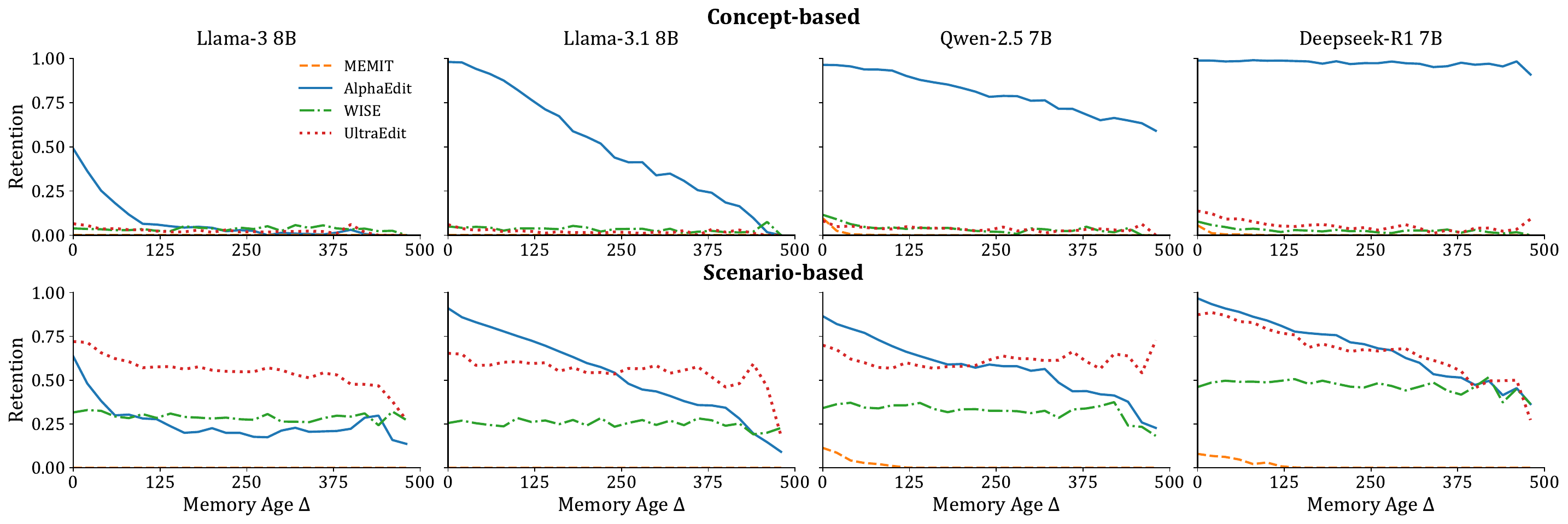}
  \caption{Forgetting curves of Concept-based and Scenario-based tests. Curves are smoothed via mean aggregation with a bin size of 10 for visual clarity. Parametric editing displays non-biological forgetting dynamics: concept-based editing often exhibits extreme dynamics, ranging from rapid memory collapse to non-decay, whereas scenario-based editing displays non-monotonic recall patterns under relational questions.
  }
  \label{fig_fc}
\end{figure*}

\subsection{Generalization and Fluency} We introduce Gen (Generalization) and Flu (Fluency) to evaluate the quality of continuous memory updates. Specifically, Gen measures the difference in accuracy between paraphrased and original queries. A strongly negative score reveals difficulty accessing edited knowledge through alternative formulations, particularly when ES and Ret remain high. Flu monitors language generation stability through the ratio of valid lexical tokens, detecting text degeneration caused by cumulative parameter modifications.

\section{Experiments}

\subsection{Experimental Setup}

\noindent \textbf{Benchmark Details.} In the dataset generation process, profiles for concept-based QA are randomly sampled from a name pool containing 2,500 distinct profiles. For scenario-based QA generation, we initialize a population of 20 agents, all instantiated using Qwen2.5-32B. These agents engaged in more than 50 rounds of interaction, resulting in an interaction graph comprising 56,095 nodes and 96,082 edges. 
We randomly sample 2,500 subgraphs from the interaction graph, each containing about 20 nodes with a maximum pairwise overlap of 20\%. These subgraphs are then provided to a dedicated QA-generation agent for question synthesis. Following this process, we constructed a benchmark consisting of 6,431 question instances in total. We provide more details in the supplementary material.

\begin{table*}[!htbp]
\small
\centering
\resizebox{0.98\textwidth}{!}{
\begin{tabular}{cc|cc|ccc|c|cc}
\toprule
\textbf{Model} & \textbf{Editing Method }
&\textbf{ ES ($\uparrow$)} &\textbf{ Ret ($\uparrow$) }
&\textbf{ $\Delta_{1/2}$ ($\uparrow$)} 
& \textbf{AUF ($\uparrow$)} &\textbf{ Decay ($\rightarrow0$)} & \textbf{TC ($\uparrow$)} 
& \textbf{Gen ($\uparrow$)} & \textbf{Flu ($\uparrow$)} \\
\midrule

\multirow{4}{*}{Llama-3 8B}
& MEMIT      & 0 & 0 & -   & -     & -      & -  & - & 0.001 \\
& WISE       & 0.02 & 0.036 & 52  & 0.035 & \underline{4e-5}   & 0.927 & \underline{\textbf{0.010}} & \underline{0.869} \\
& AlphaEdit  & \underline{0.56} & \underline{0.275} & \underline{39}  & \underline{0.076} & 1e-3  & 0.868  & -0.053 & 0.313 \\
& UltraEdit  & 0.06 & 0.032 & 14 & 0.025 & 1e-4  & \underline{0.936}  & -0.008 & 0.809 \\
\midrule

\multirow{4}{*}{Llama-3.1 8B}
& MEMIT      & 0 & 0 & -   & -     & -      & - & - & 0.001 \\
& WISE       & 0.02 & 0.031 & 82   & 0.033 & \underline{4e-5}  & 0.924  & \underline{-0.001} & 0.910 \\
& AlphaEdit  & \underline{0.98} & \underline{0.733} & \underline{236} & \underline{0.510} & 2e-3  &  0.717 & -0.161 & 0.896 \\
& UltraEdit  & 0.06 & 0.023 & 15 & 0.019 & 1e-4  & \underline{0.950}  & -0.003 & \underline{\textbf{0.923}} \\
\midrule

\multirow{4}{*}{Qwen-2.5 7B}
& MEMIT      & 0.20 & 0.059 & 5   & 0.005 & 4e-4 &  \underline{0.986}  & -0.042 & 0.054 \\
& WISE       & 0.10 & 0.037 & 25 & 0.036 & 2e-4  &  0.908 & \underline{-0.003} & 0.700 \\
& AlphaEdit  & \underline{0.92} & \underline{0.885} & \underline{452} & \underline{0.804} & 1e-3  & 0.792  & -0.559 & \underline{0.885} \\
& UltraEdit  & 0.12 & 0.044 & 12 & 0.036 & \underline{2e-4}  & 0.917 & -0.019 & 0.726 \\
\midrule

\multirow{4}{*}{Deepseek-R1 7B}
& MEMIT      & 0.12 & 0.034 & 6   & 0.003 & 2e-4  & \underline{\textbf{0.988}}   & -0.022 & 0.059 \\
& WISE       & 0.10 & 0.030 & 11 & 0.027 & 2e-4  &  0.933  & \underline{-0.010} & 0.630 \\
& AlphaEdit  & \underline{\textbf{1.00}} & \underline{\textbf{0.979}} & \underline{\textbf{>500}} &  \underline{\textbf{0.974}} & \underline{\textbf{0}}    & 0.962    & -0.930 & 0.618 \\
& UltraEdit  & 0.10 & 0.066 & 95 & 0.054 & 2e-4   & 0.870  & -0.043 & \underline{0.698} \\
\bottomrule
\end{tabular}
}
\caption{Concept-based evaluation results on ForgetBench. The best result within each model is underlined for every metric, while the overall best result is highlighted in bold face. AlphaEdit illustrates a pronounced retention-generalization trade-off: strong historical-query accuracy accompanies a substantial paraphrase penalty.}
\label{tab_concept}

\end{table*}

\noindent \textbf{Baselines and Evaluation Protocols.} We select four representative knowledge editing approaches, including MEMIT \cite{meng2022mass}, WISE \cite{wang2024wise}, AlphaEdit \cite{fang2024alphaedit}, and UltraEdit \cite{gu2025ultraedit}. We evaluate the long-term memory capabilities across four mainstream large language models, including Llama-3 8B/Llama-3.1 8B \cite{grattafiori2024llama}, Qwen-2.5 7B \cite{qwen2}, and DeepSeek-R1 7B \cite{guo2025deepseek}, covering varying architectures, pre-training scales, and reasoning capabilities. The implementation details and hyperparameter settings follow the default configurations provided by EasyEdit2 \cite{xu2025easyedit2}.

We adopt our proposed evaluation metrics, including basic statistical measures, namely ES and Ret, which measures the acquisition and retention of knowledge, and metrics associated with forgetting dynamics, including the half-life $\Delta_{1/2}$, Decay and AUF.

Preliminary experiments show that, most language models and knowledge editing methods suffer substantial degradation in both ES and Ret once the sequence length exceeds 500 in either evaluation scenario. Consequently, we uniformly set the maximum sequence length $T$ to 500 throughout all experiments. Furthermore, we set the sequence interval $k$ to 10, which provides a sufficiently fine-grained assessment of forgetting behavior while avoiding excessive computational cost. All experiments are conducted on 4 NVIDIA GeForce RTX 4090 GPUs, each with 24 GB of memory.

\subsection{Experimental Results}

\noindent \textbf{Evaluation on the Concept-based Test.} Tab.~\ref{tab_concept} exposes a pronounced retention-generalization trade-off in sequential parametric editing. AlphaEdit consistently dominates retention-oriented metrics, yet its high historical-query accuracy coincides with substantial losses under paraphrasing. On DeepSeek-R1, its Ret reaches 0.979 while Gen falls to -0.930; on Qwen-2.5, Ret reaches 0.885 with Gen at -0.559. This pattern reveals a distinctive profile of null-space-constrained editing: previously edited answers remain highly accessible through their original queries, while alternative formulations retrieve them much less reliably. 
The observed gap indicates that preservation of answers to original queries does not necessarily imply equally flexible access to the same edited knowledge.
Other editors show smaller paraphrase penalties alongside markedly weaker historical accuracy. By tracing both quantities across a long editing stream, ForgetBench exposes how durable factual retention and flexible access can diverge, a behavior obscured by immediate edit-success measurements alone. 

\begin{table*}[tbp]
\centering
\small
\resizebox{0.98\textwidth}{!}{
\begin{tabular}{cc|cc|ccc|c|cc}
\toprule
\textbf{Model} & \textbf{Editing Method }
&\textbf{ ES ($\uparrow$)} &\textbf{ Ret ($\uparrow$) }
&\textbf{ $\Delta_{1/2}$ ($\uparrow$)} 
& \textbf{AUF ($\uparrow$)} &\textbf{ Decay ($\rightarrow0$)} & \textbf{TC ($\uparrow$)} 
& \textbf{Gen ($\uparrow$)} & \textbf{Flu ($\uparrow$)} \\
\midrule

\multirow{4}{*}{Llama-3 8B}
& MEMIT      & 0 & 0 & -   & -     & -       & - & - & 0.021 \\
& WISE       & 0.30 & 0.276 & \underline{271}  & 0.290 & \underline{6e-4}    & 0.749 & \underline{-0.008} & \underline{0.996} \\
& AlphaEdit  & 0.78 & 0.369 & 47  & 0.259 & 1e-3    & \underline{0.821} & -0.047 & 0.703 \\
& UltraEdit  & \underline{0.92} & \underline{0.557} & 202 & \underline{0.553} & 1e-3    & 0.752 & -0.014 & 0.539 \\
\midrule

\multirow{4}{*}{Llama-3.1 8B}
& MEMIT      & 0 & 0 & -   & -     & -       & - & - & 0.001 \\
& WISE       & 0.30 & 0.258 & 311   & 0.252 & \underline{6e-4}    & 0.690 & \underline{-0.002} & \underline{\textbf{0.999}} \\
& AlphaEdit  & \underline{\textbf{0.96}} & \underline{0.657} & 260 & 0.539 & 1e-3   & \underline{0.932} & -0.155 & 0.801 \\
& UltraEdit  & 0.76 & 0.511 & \underline{398} & \underline{0.554} & 1e-3   & 0.727 & -0.014 & 0.882 \\
\midrule

\multirow{4}{*}{Qwen-2.5 7B}
& MEMIT      & 0.14 & 0.110 & 26   & 0.012 & 2e-4   &\underline{\textbf{0.978}} & -0.002 & 0.065 \\
& WISE       & 0.26 & 0.322 & 399 & 0.326 & 5e-4   & 0.685  & \underline{-0.001} & \underline{0.978} \\
& AlphaEdit  & \underline{\textbf{0.96}} & \underline{0.691} & 315 & 0.576 & \underline{\textbf{-8e-5}}  & 0.917 & -0.137 & 0.944 \\
& UltraEdit  & 0.74 & 0.594 & \underline{458} & \underline{0.611} & -5e-4 &  0.816 & -0.036 & 0.858 \\
\midrule

\multirow{4}{*}{Deepseek-R1 7B}
& MEMIT      & 0.12 & 0.073 & 10   & 0.012 & 2e-4  & \underline{0.968}  & -0.004 & 0.079 \\
& WISE       & 0.40 & 0.425 & \underline{\textbf{448}} & 0.468 & 8e-4   & 0.769 & \underline{\textbf{0.010}} & \underline{0.925} \\
& AlphaEdit  & \underline{0.94} & \underline{\textbf{0.784}} & 368 & \underline{\textbf{0.687}} & \underline{-1e-4} &  0.953  & -0.327 & 0.874 \\
& UltraEdit  & 0.90 & 0.755 & 403 & 0.673 & 1e-3  & 0.857  & -0.022 & 0.483 \\
\bottomrule
\end{tabular}
}
\caption{Scenario-based evaluation results on ForgetBench. Structured relational questions reshape the retention-generalization trade-off and reveal distinct long-term recall patterns across editors.
}
\label{tab_scenario}
\end{table*}

Other baselines fail to navigate this trade-off, collapsing on either side of the spectrum. Without explicit interference control, MEMIT suffers complete parametric amnesia (near-zero Ret on Llama-3 variants), although it maintains high TC due to static, un-updated prediction states. Conversely, WISE acts as a conservative baseline; it preserves generalization (Gen $\approx$ 0) and generation fluency (Flu) via its routing mechanism but shows limited access to its routed parametric updates in the concept-based setting, yielding weak long-term retention. 
UltraEdit similarly yields limited retention, indicating that strong single-step editing performance does not necessarily translate into long-term memory stability.

The forgetting curves in the upper panel of Fig.~\ref{fig_fc} further reveal that continual parametric editing can produce trajectories that differ substantially from the classic Ebbinghaus model \cite{ebbinghaus2013image}. 
While the Ebbinghaus curve is characterized by smooth monotonic decay, concept-based editing exhibits sharply different profiles: MEMIT can collapse rapidly, whereas AlphaEdit maintains high recall across the observed horizon. Across backbones, Qwen-2.5 and DeepSeek-R1 show stronger intrinsic memory robustness, showing that base model significantly influences resistance to parametric interference.

\noindent \textbf{Evaluation on the Scenario-based Test.} Tab.~\ref{tab_scenario} shows that structured relational questions substantially reshape the retention-generalization trade-off. Several editors benefit strongly from the scenario setting: UltraEdit on Llama-3 8B rises from near-collapse in the concept-based test to competitive ES and Ret, while AlphaEdit's Gen on DeepSeek-R1 improves from -0.930 to -0.327. In this setting, edited facts are embedded in relational structures, allowing the model to access an answer through connected entities and interaction patterns rather than through an isolated fact-query association. The improvement is method- and backbone-dependent: AlphaEdit's Ret changes from 0.733 to 0.657 on Llama-3.1, from 0.885 to 0.691 on Qwen-2.5, and from 0.979 to 0.784 on DeepSeek-R1. 
These results show that the balance among recent-edit accuracy, historical accuracy, and paraphrase robustness differs across the two evaluation settings. Read together, the two settings reveal complementary routes to durable recall: continual editing must preserve individual factual associations while also enabling edited knowledge to remain accessible through related entities and structured relations.


The lower panel of Fig.~\ref{fig_fc} reveals a distinct pattern of forgetting dynamics. Unlike the sharp collapse or sustained recall often observed for isolated facts, relational-query accuracy can fluctuate and even rise overall as memory age increases, as reflected by negative Decay values. Thus, subsequent edits do not affect all historical knowledge in the same way: isolated factual recall may decline rapidly, whereas relational queries can remain stable as memory age increases.


Ultimately, these two tests provide complementary views of long-term parametric memory: the concept-based test probes isolated factual retention, while the scenario-based test probes access to knowledge expressed through structured relations. This contrast highlights that while human memory integrates factual precision with associative flexibility \cite{mcclelland1995there}, current LLM editing paradigms are forced to trade one for the other.

\section{Conclusion}

In this work, we introduced ForgetBench, a benchmark for studying forgetting dynamics in the long-term parametric memory of LLMs under continual knowledge editing. Moving beyond static, single-step evaluations, ForgetBench models knowledge retention as a temporal process and provides a systematic framework for analyzing age-dependent forgetting trajectories. Our experiments across diverse models and editing methods reveal distinct memory trajectories and a recurring tension between long-term factual retention and query generalization, showing that transient editing success does not necessarily translate into stable or flexible access to edited knowledge. These findings highlight the importance of developing more robust parameter-update mechanisms that can better maintain previously acquired knowledge while adapting to new information.

\section*{Limitations} First, ForgetBench relies on synthetic or semi-synthetic datasets, which may not fully capture the complexity of real-world knowledge evolution. Future work could integrate continuously evolving corpora to enhance realism. Second, our benchmark focuses exclusively on sequential knowledge editing. Extending evaluations to alternative memory update mechanisms, such as continual pretraining or reinforcement learning-based adaptation, would provide a more comprehensive understanding of long-term memory dynamics in LLMs.

\section*{Ethical Considerations} ForgetBench is constructed primarily from synthetic profiles, simulated agent interactions, and LLM-generated question-answer pairs. It does not involve human participants, private communications, or personally identifiable user data; therefore, institutional ethics review was not required. Some generated examples may contain the names of public figures solely as synthetic entities. The associated attributes, interactions, and events should not be interpreted as factual claims about those individuals. As LLM-generated content may reflect social or linguistic biases, we recommend that the benchmark be used only for research on knowledge editing and memory evaluation. Any future extension using real-world or user-generated data should conduct additional privacy, consent, licensing, and bias assessments.

\bibliography{custom}

@article{meng2022locating,
  title={Locating and editing factual associations in gpt},
  author={Meng, Kevin and Bau, David and Andonian, Alex and Belinkov, Yonatan},
  journal={Advances in Neural Information Processing Systems},
  volume={35},
  pages={17359--17372},
  year={2022}
}

@article{mitchell2021fast,
  title={Fast model editing at scale},
  author={Mitchell, Eric and Lin, Charles and Bosselut, Antoine and Finn, Chelsea and Manning, Christopher D},
  journal={arXiv preprint arXiv:2110.11309},
  year={2021}
}

@article{meng2022mass,
  title={Mass-editing memory in a transformer},
  author={Meng, Kevin and Sharma, Arnab Sen and Andonian, Alex and Belinkov, Yonatan and Bau, David},
  journal={arXiv preprint arXiv:2210.07229},
  year={2022}
}

@article{fang2024alphaedit,
  title={Alphaedit: Null-space constrained knowledge editing for language models},
  author={Fang, Junfeng and Jiang, Houcheng and Wang, Kun and Ma, Yunshan and Jie, Shi and Wang, Xiang and He, Xiangnan and Chua, Tat-Seng},
  journal={arXiv preprint arXiv:2410.02355},
  year={2024}
}

@inproceedings{levy2017zero,
  title={Zero-shot relation extraction via reading comprehension},
  author={Levy, Omer and Seo, Minjoon and Choi, Eunsol and Zettlemoyer, Luke},
  booktitle={Proceedings of the 21st Conference on Computational Natural Language Learning (CoNLL 2017)},
  pages={333--342},
  year={2017}
}

@article{deng2024everything,
  title={Everything is editable: Extend knowledge editing to unstructured data in large language models},
  author={Deng, Jingcheng and Wei, Zihao and Pang, Liang and Ding, Hanxing and Shen, Huawei and Cheng, Xueqi},
  journal={arXiv preprint arXiv:2405.15349},
  year={2024}
}

@article{li2025reinforced,
  title={Reinforced lifelong editing for language models},
  author={Li, Zherui and Jiang, Houcheng and Chen, Hao and Bi, Baolong and Zhou, Zhenhong and Sun, Fei and Fang, Junfeng and Wang, Xiang},
  journal={arXiv preprint arXiv:2502.05759},
  year={2025}
}

@article{gu2026hierarchical,
  title={Hierarchical Orthogonal Residual Spread for Precise Massive Editing in Large Language Models},
  author={Gu, Xiaojie and Chen, Guangxu and Yang, Yuheng and Han, Jingxin and Zhang, Andi},
  journal={arXiv preprint arXiv:2601.11441},
  year={2026}
}

@article{tan2023massive,
  title={Massive editing for large language models via meta learning},
  author={Tan, Chenmien and Zhang, Ge and Fu, Jie},
  journal={arXiv preprint arXiv:2311.04661},
  year={2023}
}

@article{pan2025precise,
  title={Precise localization of memories: A fine-grained neuron-level knowledge editing technique for llms},
  author={Pan, Haowen and Wang, Xiaozhi and Cao, Yixin and Shi, Zenglin and Yang, Xun and Li, Juanzi and Wang, Meng},
  journal={arXiv preprint arXiv:2503.01090},
  year={2025}
}

@incollection{atkinson1968human,
  title={Human memory: A proposed system and its control processes},
  author={Atkinson, Richard C and Shiffrin, Richard M},
  booktitle={Psychology of Learning and Motivation},
  volume={2},
  pages={89--195},
  year={1968},
  publisher={Elsevier}
}

@article{lewis2020retrieval,
  title={Retrieval-augmented generation for knowledge-intensive nlp tasks},
  author={Lewis, Patrick and Perez, Ethan and Piktus, Aleksandra and Petroni, Fabio and Karpukhin, Vladimir and Goyal, Naman and K{\"u}ttler, Heinrich and Lewis, Mike and Yih, Wen-tau and Rockt{\"a}schel, Tim and others},
  journal={Advances in Neural Information Processing Systems},
  volume={33},
  pages={9459--9474},
  year={2020}
}

@inproceedings{asai2023self,
  title={Self-rag: Learning to retrieve, generate, and critique through self-reflection},
  author={Asai, Akari and Wu, Zeqiu and Wang, Yizhong and Sil, Avirup and Hajishirzi, Hannaneh},
  booktitle={The Twelfth International Conference on Learning Representations},
  year={2023}
}

@inproceedings{chevalier2023adapting,
  title={Adapting language models to compress contexts},
  author={Chevalier, Alexis and Wettig, Alexander and Ajith, Anirudh and Chen, Danqi},
  booktitle={Proceedings of the 2023 Conference on Empirical Methods in Natural Language Processing},
  pages={3829--3846},
  year={2023}
}

@article{li2024snapkv,
  title={Snapkv: Llm knows what you are looking for before generation},
  author={Li, Yuhong and Huang, Yingbing and Yang, Bowen and Venkitesh, Bharat and Locatelli, Acyr and Ye, Hanchen and Cai, Tianle and Lewis, Patrick and Chen, Deming},
  journal={Advances in Neural Information Processing Systems},
  volume={37},
  pages={22947--22970},
  year={2024}
}

@article{wei2022chain,
  title={Chain-of-thought prompting elicits reasoning in large language models},
  author={Wei, Jason and Wang, Xuezhi and Schuurmans, Dale and Bosma, Maarten and Xia, Fei and Chi, Ed and Le, Quoc V and Zhou, Denny and others},
  journal={Advances in Neural Information Processing Systems},
  volume={35},
  pages={24824--24837},
  year={2022}
}

@inproceedings{xu2025softcot,
  title={Softcot: Soft chain-of-thought for efficient reasoning with llms},
  author={Xu, Yige and Guo, Xu and Zeng, Zhiwei and Miao, Chunyan},
  booktitle={Proceedings of the 63rd Annual Meeting of the Association for Computational Linguistics (Volume 1: Long Papers)},
  pages={23336--23351},
  year={2025}
}

@inproceedings{yang2018hotpotqa,
  title={HotpotQA: A dataset for diverse, explainable multi-hop question answering},
  author={Yang, Zhilin and Qi, Peng and Zhang, Saizheng and Bengio, Yoshua and Cohen, William and Salakhutdinov, Ruslan and Manning, Christopher D},
  booktitle={Proceedings of the 2018 Conference on Empirical Methods in Natural Language Processing},
  pages={2369--2380},
  year={2018}
}

@inproceedings{bai2024longbench,
  title={Longbench: A bilingual, multitask benchmark for long context understanding},
  author={Bai, Yushi and Lv, Xin and Zhang, Jiajie and Lyu, Hongchang and Tang, Jiankai and Huang, Zhidian and Du, Zhengxiao and Liu, Xiao and Zeng, Aohan and Hou, Lei and others},
  booktitle={Proceedings of the 62nd Annual Meeting of the Association for Computational Linguistics (volume 1: Long papers)},
  pages={3119--3137},
  year={2024}
}

@inproceedings{maharana2024evaluating,
  title={Evaluating very long-term conversational memory of llm agents},
  author={Maharana, Adyasha and Lee, Dong-Ho and Tulyakov, Sergey and Bansal, Mohit and Barbieri, Francesco and Fang, Yuwei},
  booktitle={Proceedings of the 62nd Annual Meeting of the Association for Computational Linguistics (Volume 1: Long Papers)},
  pages={13851--13870},
  year={2024}
}

@article{ebbinghaus2013image,
  title={Memory: A contribution to experimental psychology},
  author={Ebbinghaus, Hermann},
  journal={Annals of Neurosciences},
  volume={20},
  number={4},
  pages={155},
  year={2013}
}

@article{gu2025ultraedit,
  title={UltraEdit: Training-, Subject-, and Memory-Free Lifelong Editing in Language Models},
  author={Gu, Xiaojie and Huang, Ziying and Gu, Jia-Chen and Zhang, Kai},
  journal={arXiv preprint arXiv:2505.14679},
  year={2025}
}

@article{wang2024wise,
  title={Wise: Rethinking the knowledge memory for lifelong model editing of large language models},
  author={Wang, Peng and Li, Zexi and Zhang, Ningyu and Xu, Ziwen and Yao, Yunzhi and Jiang, Yong and Xie, Pengjun and Huang, Fei and Chen, Huajun},
  journal={Advances in Neural Information Processing Systems},
  volume={37},
  pages={53764--53797},
  year={2024}
}

@article{grattafiori2024llama,
  title={The llama 3 herd of models},
  author={Grattafiori, Aaron and Dubey, Abhimanyu and Jauhri, Abhinav and Pandey, Abhinav and Kadian, Abhishek and Al-Dahle, Ahmad and Letman, Aiesha and Mathur, Akhil and Schelten, Alan and Vaughan, Alex and others},
  journal={arXiv preprint arXiv:2407.21783},
  year={2024}
}

@article{guo2025deepseek,
  title={Deepseek-r1: Incentivizing reasoning capability in llms via reinforcement learning},
  author={Guo, Daya and Yang, Dejian and Zhang, Haowei and Song, Junxiao and Wang, Peiyi and Zhu, Qihao and Xu, Runxin and Zhang, Ruoyu and Ma, Shirong and Bi, Xiao and others},
  journal={arXiv preprint arXiv:2501.12948},
  year={2025}
}

@article{qwen2,
  title={Qwen2 technical report},
  author={Yang, An and Yang, Baosong and Hui, Binyuan and Zheng, Bo and Yu, Bowen and Zhou, Chang and Li, Chengpeng and Li, Chengyuan and Liu, Dayiheng and Huang, Fei and others},
  journal={arXiv preprint arXiv:2407.10671},
  year={2024}
}

@inproceedings{liu2024forgetting,
  title={Forgetting curve: A reliable method for evaluating memorization capability for long-context models},
  author={Liu, Xinyu and Zhao, Runsong and Huang, Pengcheng and Xiao, Chunyang and Li, Bei and Wang, Jingang and Xiao, Tong and Zhu, Jingbo},
  booktitle={Proceedings of the 2024 Conference on Empirical Methods in Natural Language Processing},
  pages={4667--4682},
  year={2024}
}

@article{wu2024longmemeval,
  title={Longmemeval: Benchmarking chat assistants on long-term interactive memory},
  author={Wu, Di and Wang, Hongwei and Yu, Wenhao and Zhang, Yuwei and Chang, Kai-Wei and Yu, Dong},
  journal={arXiv preprint arXiv:2410.10813},
  year={2024}
}

@article{hu2026evermembench,
  title={EverMemBench: Benchmarking Long-Term Interactive Memory in Large Language Models},
  author={Hu, Chuanrui and Li, Tong and Gao, Xingze and Chen, Hongda and Xu, Dannong and Bai, Yi and Lin, Tianwei and Zhao, Xinda and Li, Xiaohong and An, Jiaqi and others},
  journal={arXiv preprint arXiv:2602.01313},
  year={2026}
}

@article{ding2026memground,
  title={MemGround: Long-Term Memory Evaluation Kit for Large Language Models in Gamified Scenarios},
  author={Ding, Yihang and Xia, Wanke and Zhao, Yiting and Su, Jinbo and Yang, Jialiang and Zhang, Zhengbo and Wang, Ke and Yang, Wenming},
  journal={arXiv preprint arXiv:2604.14158},
  year={2026}
}

@inproceedings{uddin2026recall,
  title={From recall to forgetting: Benchmarking long-term memory for personalized agents},
  author={Uddin, Md Nayem and Shubham, Kumar and Blanco, Eduardo and Baral, Chitta and Wang, Gengyu},
  booktitle={Findings of the Association for Computational Linguistics: ACL 2026},
  pages={26814--26841},
  year={2026}
}

@inproceedings{xu2025easyedit2,
  title={Easyedit2: An easy-to-use steering framework for editing large language models},
  author={Xu, Ziwen and Wang, Shuxun and Xu, Kewei and Xu, Haoming and Wang, Mengru and Deng, Xinle and Yao, Yunzhi and Zheng, Guozhou and Chen, Huajun and Zhang, Ningyu},
  booktitle={Proceedings of the 2025 Conference on Empirical Methods in Natural Language Processing: System Demonstrations},
  pages={522--535},
  year={2025}
}

@article{mcclelland1995there,
  title={Why there are complementary learning systems in the hippocampus and neocortex: insights from the successes and failures of connectionist models of learning and memory.},
  author={McClelland, James L and McNaughton, Bruce L and O'Reilly, Randall C},
  journal={Psychological review},
  volume={102},
  number={3},
  pages={419},
  year={1995},
  publisher={American Psychological Association}
}

@misc{thede2025wikibigeditunderstandinglimitslifelong,
      title={WikiBigEdit: Understanding the Limits of Lifelong Knowledge Editing in LLMs}, 
      author={Lukas Thede and Karsten Roth and Matthias Bethge and Zeynep Akata and Tom Hartvigsen},
      year={2025},
      eprint={2503.05683},
      archivePrefix={arXiv},
      primaryClass={cs.CL},
      url={https://arxiv.org/abs/2503.05683}, 
}

@misc{jung2026scalablelifelongknowledgeediting,
      title={Towards Scalable Lifelong Knowledge Editing with Selective Knowledge Suppression}, 
      author={Dahyun Jung and Jaewook Lee and Heuiseok Lim},
      year={2026},
      eprint={2604.19089},
      archivePrefix={arXiv},
      primaryClass={cs.AI},
      url={https://arxiv.org/abs/2604.19089}, 
}

@article{LIU2026123848,
title = {LPD-Edit: Latent principal denoising for robust lifelong knowledge editing},
journal = {Information Sciences},
volume = {756},
pages = {123848},
year = {2026},
issn = {0020-0255},
doi = {https://doi.org/10.1016/j.ins.2026.123848},
url = {https://www.sciencedirect.com/science/article/pii/S0020025526007796},
author = {Ruinan Liu and Fulong Lin and Xin Wang and Zhihua Fu}
}

\appendix

\section{Overview}

In the appendices, we provide additional details and results to support the main submission. 

Specifically, 
\begin{itemize}
    \item Section \ref{secA} presents more details of ForgetBench.
    \item Section \ref{secB} presents more evaluation results on Concept-based and Scenario-based settings.
\end{itemize}

\section{ForgetBench Details}
\label{secA}

ForgetBench is constructed to evaluate long-term memory dynamics under continual knowledge editing. It contains two complementary evaluation paradigms, namely concept-based QA and scenario-based QA, which are designed to disentangle isolated factual updates from structured relational reasoning in contextual environments.

\subsection{Algorithm Pseudocode for Scenario-based QA Generation}

We generate scenario-based QA sequences through a two-stage process. First, we simulate multiple rounds of agent interactions to construct a dynamic knowledge graph containing agent and item nodes. Second, we sample diverse subgraphs from the resulting graph and use an LLM to convert each subgraph into natural-language context and generate multiple-choice questions with sampled distractors. The complete procedure is summarized in Algo.~\ref{alg:kg_qa_generation}.

\subsection{Evaluation Metric Details}

\noindent \textbf{Smoothing of Forgetting Curves.}
To improve robustness against local fluctuations in $\mathcal{F}(\Delta)$, we apply a lightweight smoothing preprocessing step to reduce small-scale noise. Specifically, after computing the Forgetting Curve at each discrete memory age $\Delta$, we divide the curve into consecutive non-overlapping windows of 10 memory ages. For the $j$-th window, covering $\Delta\in\{10j,\ldots,10j+9\}$, we compute the arithmetic mean of all available Forgetting Curve values:
\begin{equation}
{\mathcal{F}}_j
=
\frac{1}{|\mathcal{B}_j|}
\sum_{\Delta\in\mathcal{B}_j}
\mathcal{F}(\Delta),
\mathcal{B}_j
=
\{10j,\ldots,10j+9\}.
\end{equation}
This produces a smoothed curve with one value for every 10 consecutive memory ages. The resulting smoothed curve is used to reduce local fluctuations when reporting the forgetting trajectory and estimating the memory decay trend.

\begin{algorithm}[H]
\caption{Scenario-based QA Generation}
\label{alg:kg_qa_generation}

\textbf{Input}: Simulation rounds $K$, subgraph size $L$, number of subgraphs $N$, questions per subgraph $t$ \\
\textbf{Output}: QA sequence $\mathcal{D}$

\begin{algorithmic}[1]

\STATE Initialize QA sequence $\mathcal{D}\leftarrow\emptyset$ and the length $T=N\times t$
\STATE \textbf{// Phase 1: KG Generation}
\STATE Initialize agent set $\mathcal{A}$, item set $\mathcal{I}$, and edge set $\mathcal{E}$

\FOR{$k=1$ to $K$}
    \STATE Update $\mathcal{A}$ by adding and deleting specific agents
    \STATE Select active agents $\mathcal{A}_{active}\subseteq\mathcal{A}$
    
    \FORALL{$a_i\in\mathcal{A}_{active}$}
        \STATE Form a collaborative group $\mathcal{A}_{col}\subseteq\mathcal{A}$ to generate item nodes and agent-item interactions   
        \STATE Update $\mathcal{I}$ and $\mathcal{E}$
    \ENDFOR
\ENDFOR 

\STATE Construct knowledge graph $\mathcal{G}=\{V=\mathcal{A}\cup\mathcal{I},E=\mathcal{E}\}$

\STATE \textbf{// Phase 2: QA Generation}
\STATE Initialize multi-choice QA generation LLM $\mathcal{F}$
\STATE Sample $N$ diverse subgraphs $\{\mathcal{G}_i\}_{i=1}^{N}$ of size $L$ from $\mathcal{G}$

\FORALL{$\mathcal{G}_i$}
    \STATE Rephrase $\mathcal{G}_i$ into natural-language context $C_i$
    
    \FOR{$j=1$ to $t$}
        \STATE $\{Q_{ij}, A_{ij}^{right}, A_1, A_2, A_3\} \leftarrow \mathcal{F}(C_i),$ $A_1$, $A_2$, $ A_3$ are incorrect
        \STATE Sample distractor answer $A_{ij}^{wrong} \sim \{A_1, A_2, A_3\}$
        \STATE Add $(Q_{ij}, A_{ij}^{right}, A_{ij}^{wrong})$ into $\mathcal{D}$
    \ENDFOR
    
\ENDFOR 

\STATE \textbf{return} $\mathcal{D}$

\end{algorithmic}
\end{algorithm}

\noindent \textbf{Generalization.}
Gen measures whether the updated memory generalizes beyond the exact wording of the original query. For each knowledge item, we evaluate the edited model on both the original question and a semantically equivalent paraphrased question. We define Gen as the difference between the overall accuracy on paraphrased queries and that on original queries:
\begin{equation}
\text{Gen}
=
\text{Acc}_{\text{para}}
-
\text{Acc}_{\text{orig}},
\end{equation}
where $\text{Acc}_{\text{para}}$ and $\text{Acc}_{\text{orig}}$ denote the overall accuracy on all evaluated paraphrased and original queries, respectively.

To construct the paraphrased queries, we use Qwen2.5-32B to rewrite each original multiple-choice question. The model is instructed to rewrite the question in a natural way while preserving exactly the same meaning. Specifically, the prompt requires that the answer options, option labels, and answer meaning remain unchanged, and that only the rewritten multiple-choice question be output. Since the original and paraphrased questions share the same answer choices and ground-truth answer, the resulting accuracy difference evaluates the robustness of the updated memory to changes in query formulation.

\noindent \textbf{Fluency.} Flu measures the linguistic fluency of generated responses as the proportion of valid lexical tokens. Given a generated response, we first lowercase the text and split it by whitespace. We then remove leading and trailing punctuation from each token and discard empty tokens. A token is considered valid if it appears in the NLTK English word vocabulary (\texttt{nltk.corpus.words}), augmented with the answer-choice tokens ${a,b,c,d}$ and numerical tokens. Formally,
\begin{equation}
\text{Flu}(y)=
\frac{\sum_{w\in\mathcal{T}(y)}\mathbb{I}[w\in\mathcal{V}]}
{|\mathcal{T}(y)|},
\end{equation}
where $\mathcal{T}(y)$ denotes the resulting non-empty tokens and $\mathcal{V}$ is the augmented vocabulary. If no valid tokens remain after preprocessing, Flu is defined as $0$. The final Flu score is the average Flu over all evaluated responses.

\subsection{Dataset Examples}

\noindent \textbf{Concept-based QA.} The concept-based setting focuses on atomic factual modifications, where each query targets a single independent fact. Each sample consists of a question $Q$, an original answer $A^{old}$, and an updated answer $A^{new}$ (a rephrased variant $Q'$ for generalization test). Several examples are listed below.

\begin{tcolorbox}[
  colback=selfevolagent_light!20,
  colframe=selfevolagent_light!80,
  colbacktitle=selfevolagent_light!80,
  coltitle=white,
  breakable,
  title={\bfseries{Concept-based QA Instances}},
  boxrule=2pt,
  arc=5pt,
  drop shadow,
  before skip=15pt,
  after skip=15pt,
  left=5pt,   
  right=5pt,
  fontupper=\small
]
\begin{itemize}[leftmargin=*]
\item Item 1:
    \begin{itemize}
        \item \textbf{Question $Q$}: How old is Benjamin White?
        \item \textbf{Rephrased Question $Q'$}: What is Benjamin White's age?
        \item \textbf{Former Answer $A^{old}$}: 50
        \item \textbf{Updated Answer $A^{new}$}: 37
    \end{itemize}
\item Item 2:
    \begin{itemize}
        \item \textbf{Question $Q$}: How old is Mike Rivera?
        \item \textbf{Rephrased Question $Q'$}: What is Mike Rivera's age?
        \item \textbf{Former Answer $A^{old}$}: 73
        \item \textbf{Updated Answer $A^{new}$}: 60
    \end{itemize}
\end{itemize}
\label{example1}
\end{tcolorbox}

\noindent \textbf{Scenario-based QA.} The scenario-based setting constructs temporally ordered interaction or event-driven contexts, where knowledge is embedded in realistic social media. Each sample consists of a context $C$ describing evolving interactions over time. This setting evaluates whether models can maintain consistent reasoning over interconnected knowledge under continual updates. A few illustrative instances of $C$ are provided below.

\begin{figure*}[bt]
  \centering
    \centering
    \includegraphics[width=\linewidth]{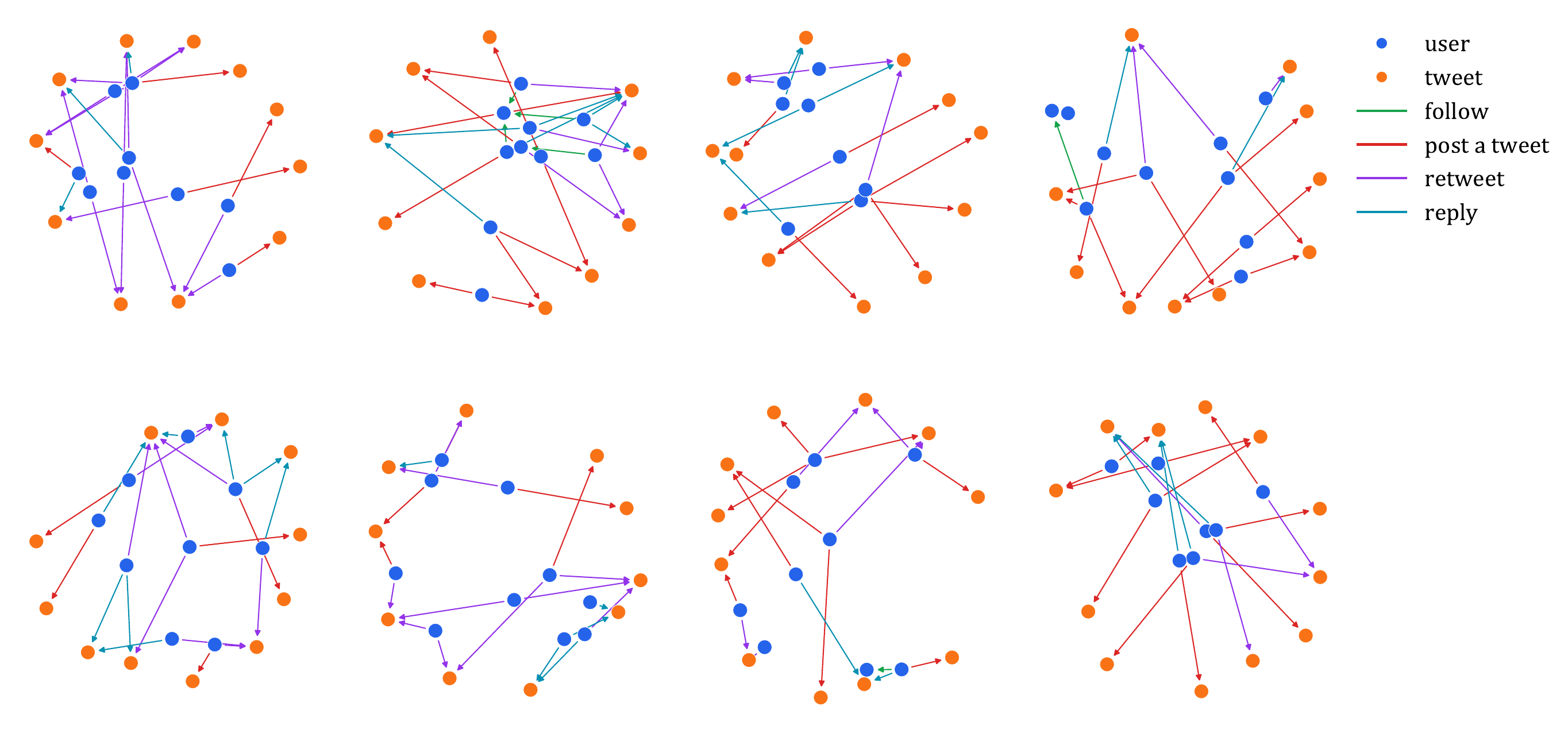}
  \caption{Subgraph samples in Scenario-based QA generation.}
  \label{fig_graph}
\end{figure*}

\begin{tcolorbox}[
  colback=selfevolagent_light!20,
  colframe=selfevolagent_light!80,
  colbacktitle=selfevolagent_light!80,
  coltitle=white,
  title={\bfseries{Scenario-based Context Instances}},
  boxrule=2pt,
  arc=5pt,
  drop shadow,
  breakable,
  before skip=15pt,
  after skip=15pt,
  left=5pt,   
  right=5pt,
  fontupper=\small
]
\begin{itemize}[leftmargin=*]
\item Item 1:
    \begin{itemize}
        \item  \textbf{Context $C$:} On 2024-06-04, NationalGeographic replied to a tweet, focusing on AI and space exploration, commenting: "Looking forward to integrating AI with space missions to enhance autonomy and efficiency. This could lead to groundbreaking discoveries and pave the way for human exploration of distant worlds. \#AIInSpace \#FutureOfExploration".
    \end{itemize}
\item Item 2:
    \begin{itemize}
        \item  \textbf{Context $C$:   }  
        On 2024-04-24, Beyonce followed the user shinebrightart.
        On 2024-05-17, Beyonce followed the user TheEllenShow.
        On 2024-05-20, Beyonce followed the user KatyPerry. 
        On 2024-05-22, Beyonce followed the user Beyonce.
        On 2024-05-26, Beyonce followed the user LadyGaga.
        On 2024-05-27, Beyonce followed the user consequence.
        On 2024-06-04, Beyonce followed the user jtimberlake.
    \end{itemize}
\end{itemize}
\end{tcolorbox}

Based on the constructed contexts, we further build multiple-choice QA tasks that require models to retrieve correct temporal or causal information. Fig.~\ref{fig_graph} illustrates representative subgraph samples used in scenario-based QA generation. These graphs encode structured interactions among users, posts, and actions (e.g., follow, tweet, reply), providing a unified view for constructing temporally grounded QA tasks. Below are the QA instances corresponding to the former contexts:

\begin{tcolorbox}[
  colback=selfevolagent_light!20,
  colframe=selfevolagent_light!80,
  colbacktitle=selfevolagent_light!80,
  coltitle=white,
  title={\bfseries{Scenario-based QA Instances}},
  boxrule=2pt,
  arc=5pt,
  drop shadow,
  breakable,
  before skip=15pt,
  after skip=15pt,
  left=5pt,   
  right=5pt,
  fontupper=\small
]
\begin{itemize}[leftmargin=*]
\item Item 1:
    \begin{itemize}
        \item \textbf{Question $Q$}: Which of the following entities has discussed the integration of AI with space missions?
Options:
    \begin{itemize}
        \item A. NatureLover89
        \item B. NYTimes
        \item C. International Cooperation in AI for Public Health
        \item D. NationalGeographic
    \end{itemize}
        \item \textbf{Rephrased Question $Q'$}: Which entity below has talked about combining AI with space missions?
Options:
    \begin{itemize}
        \item A. NatureLover89
        \item B. NYTimes
        \item C. International Cooperation in AI for Public Health
        \item D. NationalGeographic
    \end{itemize}
        \item \textbf{Former Answer $A^{old}$}: B
        \item \textbf{Updated Answer $A^{new}$}: D
    \end{itemize}
\item Item 2:
    \begin{itemize}
        \item \textbf{Question $Q$}: On which date did Beyoncé follow KatyPerry on social media?
Options:
    \begin{itemize}
        \item A. 2024-05-30
        \item B. 2024-05-25
        \item C. 2024-05-20
        \item D. 2024-05-15
    \end{itemize}
        \item \textbf{Rephrased Question $Q'$}: When did Beyoncé start following Katy Perry on social media?
Options:
    \begin{itemize}
        \item A. 2024-05-30
        \item B. 2024-05-25
        \item C. 2024-05-20
        \item D. 2024-05-15
    \end{itemize}
        \item \textbf{Former Answer $A^{old}$}: B
        \item \textbf{Updated Answer $A^{new}$}: C
    \end{itemize}
\end{itemize}
\end{tcolorbox}

\section{Additional Results}
\label{secB}

\subsection{Abnormal Model Outputs}

To further illustrate the failure cases captured by the Flu metric, we provide representative examples of abnormal model outputs after continual knowledge editing. These cases mainly correspond to response collapse, where the model generates meaningless symbols, corrupted token sequences, or incomplete responses instead of coherent answers. Such behaviors indicate severe degradation of generation reliability beyond factual forgetting.

\begin{tcolorbox}[
  colback=selfevolagent_light!20,
  colframe=selfevolagent_light!80,
  colbacktitle=selfevolagent_light!80,
  coltitle=white,
  breakable,
  title={\bfseries{Examples of Abnormal Model Outputs}},
  boxrule=2pt,
  arc=5pt,
  drop shadow,
  before skip=15pt,
  after skip=15pt,
  left=5pt,   
  right=5pt,
  fontupper=\small
]
\begin{itemize}[leftmargin=*]
\item Item 1:
    \begin{itemize}
        \item \textbf{Question $Q$}: How old is Mike Rivera?
        \item \textbf{Model Output}: ,,,,,,,,,,
    \end{itemize}
    \begin{itemize}
        \item \textbf{Question $Q$}: What type of art did Bill Gates humorously compare his tweets to?
Options:
\begin{itemize}
    \item A. Impressionism
    \item B. Abstract Art
    \item C. Surrealism
    \item D. Realism
\end{itemize}
        \item \textbf{Model Output}: ?>>;\};
        
/Internal;\};

/

;\};

/ay?>>;\};

;\};

    \end{itemize}
\end{itemize}
\label{example1}
\end{tcolorbox}

\subsection{All Evaluation Matrices}

\begin{figure*}[bt]
  \centering
    \centering
    \includegraphics[width=\linewidth]{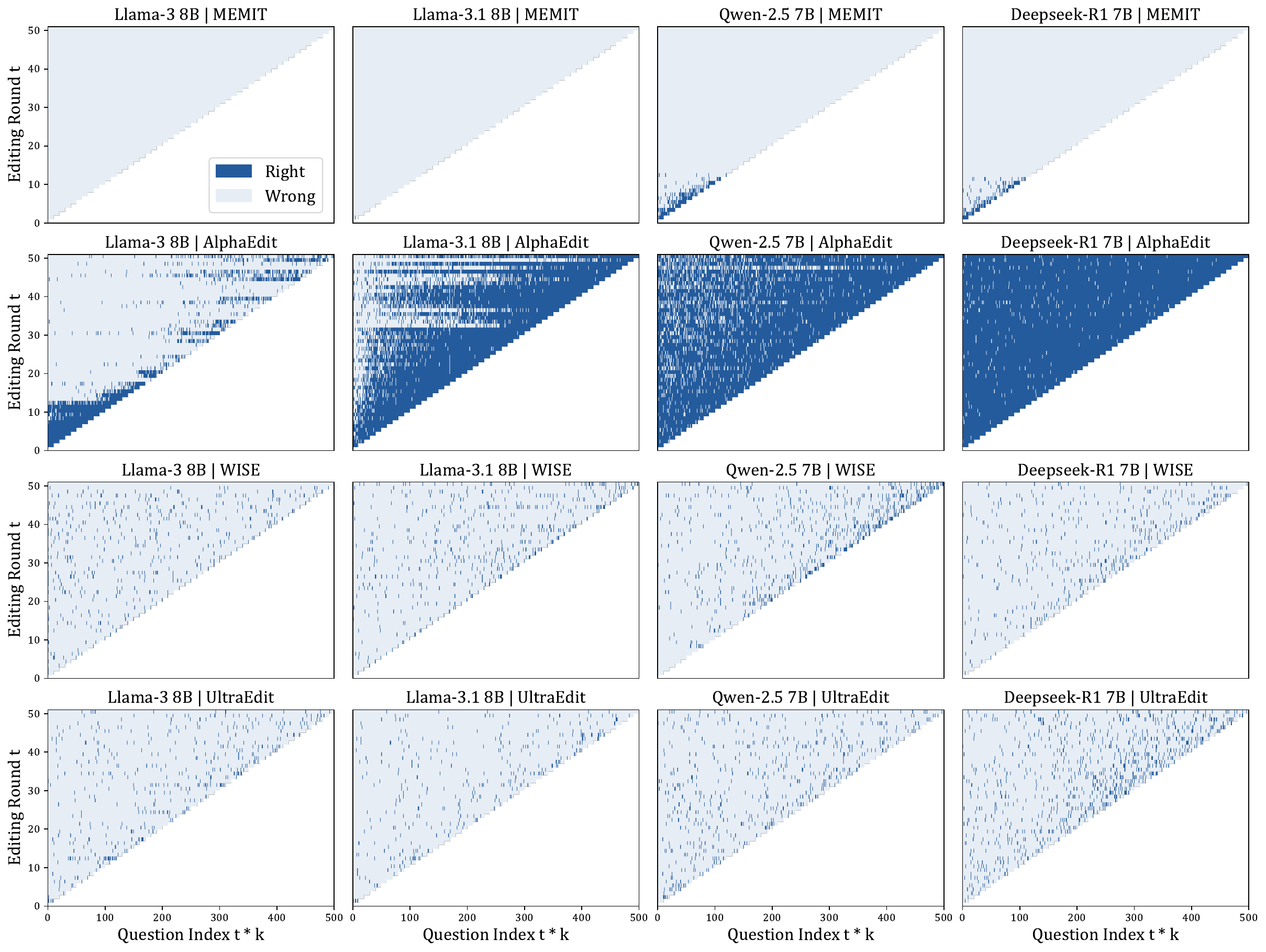}
  \caption{All $\mathcal{Z}$ matrices of Concept-based test.}
  \label{fig_concept}
\end{figure*}

\begin{figure*}[bt]
  \centering
    \centering
    \includegraphics[width=\linewidth]{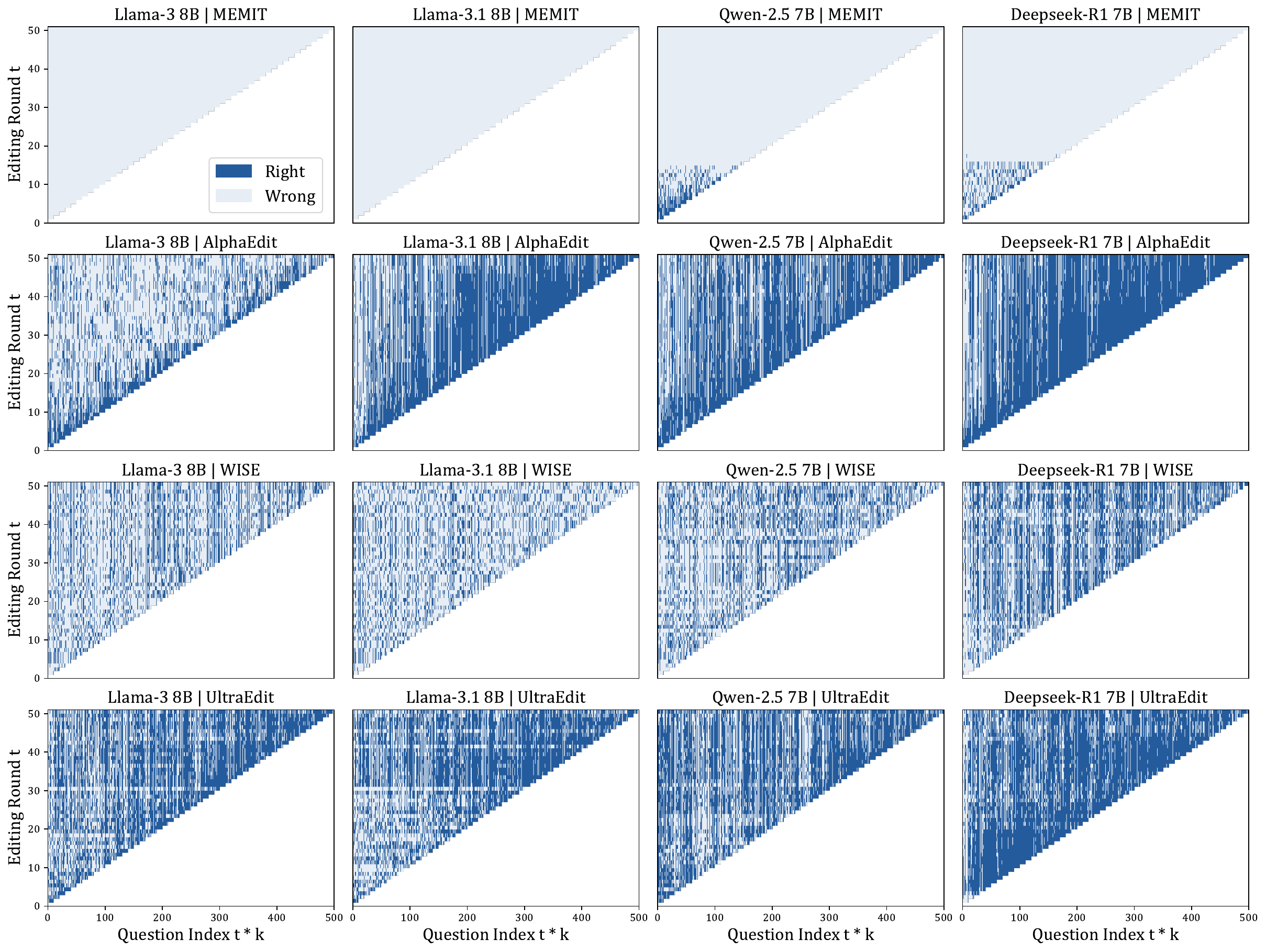}
  \caption{All $\mathcal{Z}$ matrices of Scenario-based test.}
  \label{fig_scenario}
\end{figure*}

\begin{figure*}[bt]
  \centering
    \centering
    \includegraphics[width=\linewidth]{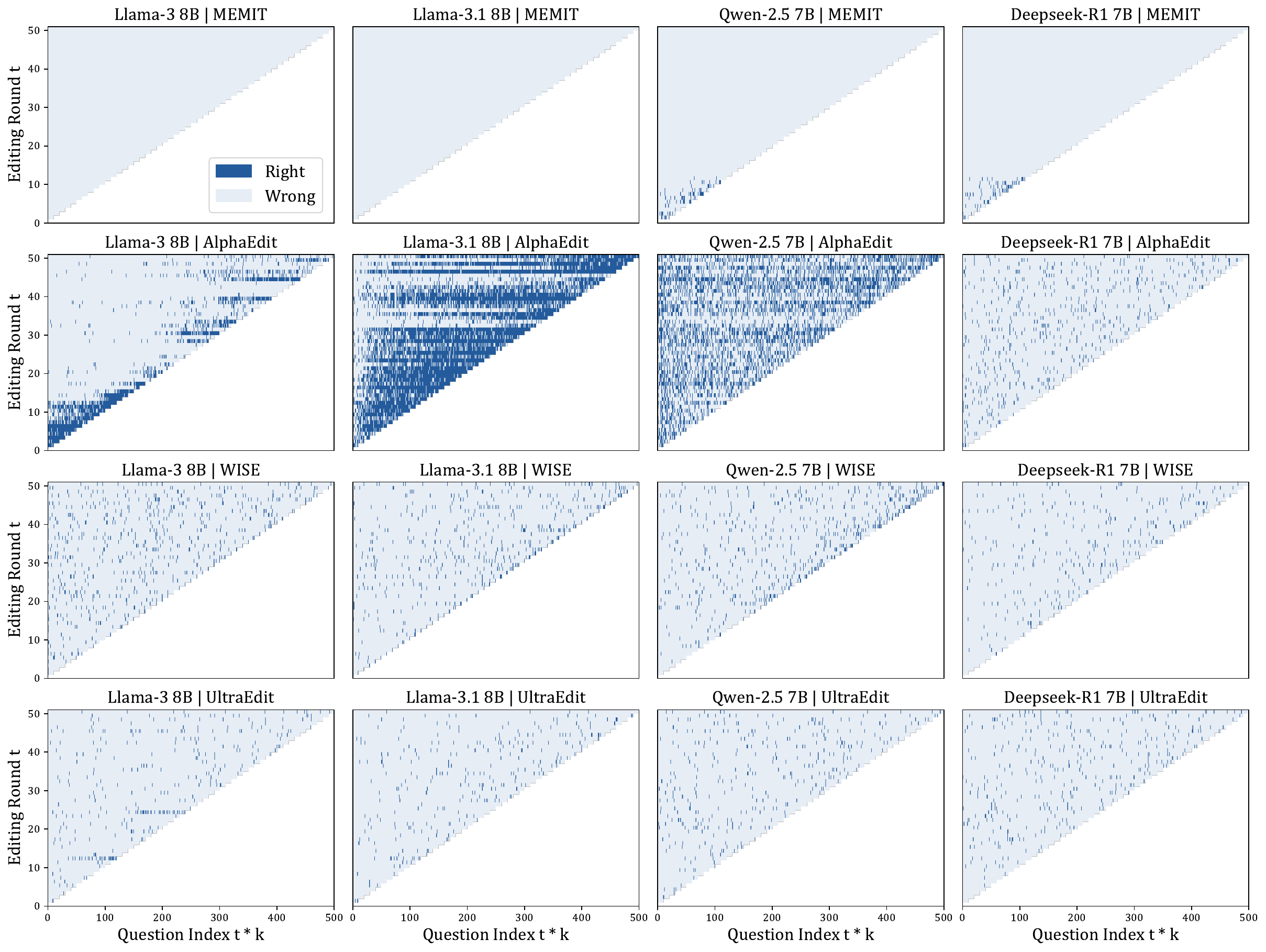}
  \caption{All $\mathcal{Z}$ matrices of Concept-based test with rephrased questions.}
  \label{fig_concept_reph}
\end{figure*}

\begin{figure*}[bt]
  \centering
    \centering
    \includegraphics[width=\linewidth]{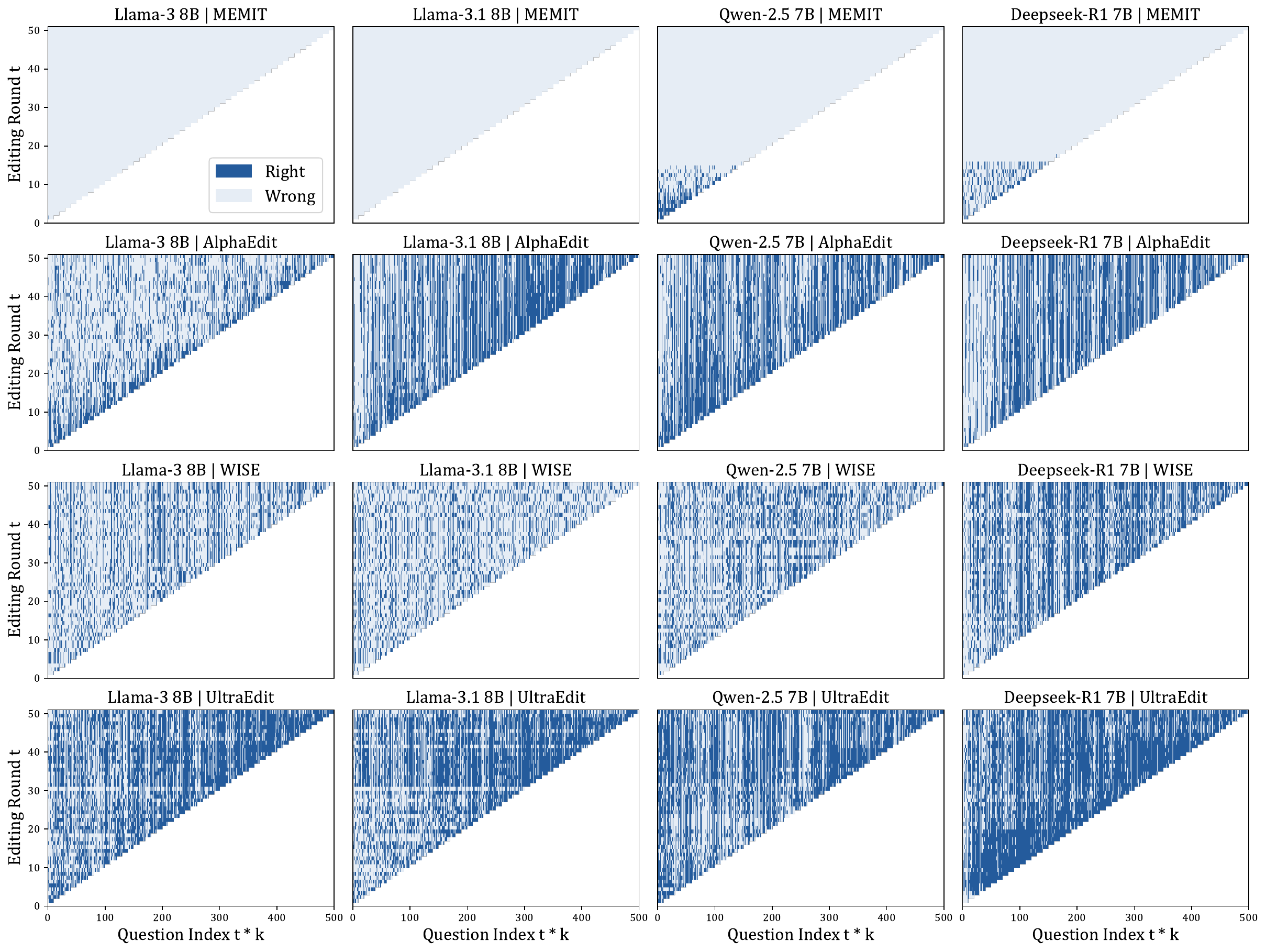}
  \caption{All $\mathcal{Z}$ matrices of Scenario-based test with rephrased questions.}
  \label{fig_scenario_reph}
\end{figure*}

We present a comprehensive set of results to further analyze the temporal behavior of different models and knowledge editing methods under the proposed ForgetBench benchmark. Fig.~\ref{fig_concept} and Fig.~\ref{fig_scenario} visualize the full evaluation matrices $\mathcal{Z}$ for concept-based and scenario-based settings, respectively. Fig.~\ref{fig_concept_reph} and Fig.~\ref{fig_scenario_reph} further present results under rephrased questions. We observe that while absolute accuracy decreases under paraphrasing, the overall temporal structure of the $\mathcal{Z}$ matrices remains consistent. This indicates that the observed forgetting behavior is not driven by surface-level lexical cues but reflects more intrinsic memory dynamics.

\end{document}